\documentclass[10pt,twocolumn,letterpaper]{article}

\usepackage{iccv}
\usepackage{times}
\usepackage{epsfig}
\usepackage{graphicx}
\usepackage{amsmath}
\usepackage{amssymb}
\usepackage{url}

\usepackage[breaklinks=true,bookmarks=false]{hyperref}

\iccvfinalcopy % *** Uncomment this line for the final submission

\ificcvfinal\fi

\begin{document}

%%%%%%%%% TITLE
\title{2nd Place Solution in Google AI Open Images Object Detection Track 2019}

\author{Ruoyu Guo, Cheng Cui, Yuning Du, Xianglong Meng, \\
Xiaodi Wang, Jingwei Liu, Jianfeng Zhu, Yuan Feng, Shumin Han\\
Baidu Inc.\\
{\tt\small \{guoruoyu, duyuning\}@baidu.com}
}

%% omit the following lines up until the closing ``}''.
%% Additional authors and addresses can be added with ``\and'',
%% just like the second author.
%% To save space, use either the email address or home page, not both
%\and

\maketitle
\ificcvfinal\thispagestyle{empty}\fi

%%%%%%%%% ABSTRACT
\begin{abstract}
   We present an object detection framework based on PaddlePaddle. We put all the strategies together (multi-scale training, FPN, Cascade, Dcnv2, Non-local, libra loss) based on ResNet200-vd backbone. Our model score on public leaderboard comes to 0.6269 with single scale test. We proposed a new voting method called top-k voting-nms, based on the SoftNMS detection results. The voting method helps us merge all the models' results more easily and achieve 2nd place in the Google AI Open Images Object Detection Track 2019.
\end{abstract}

%%%%%%%%% BODY TEXT
\section{Introduction}

Open Images Detection Dataset V5(OIDV5) is the largest existing object detection dataset with object location annotations. It contains over 1.7M images with 500 classes and over 12M annotated boxes, which is far more than current famous datasets, such as, COCO ~\cite{noauthor_coco_nodate} and Objects365 dataset ~\cite{noauthor_objects365_nodate}.

Besides large-scale and closing to real scene, OIDV5 has other typical characteristics:

The annotations of instances in OIDV5 are not overall, and for each image, instances of classes not verified to exist in the image are not annotated. It leads to treating a bounding box as the background when an unverified instance is inside the box. 

\begin{figure}[htbp]
 \centering
 \includegraphics[width=7cm]{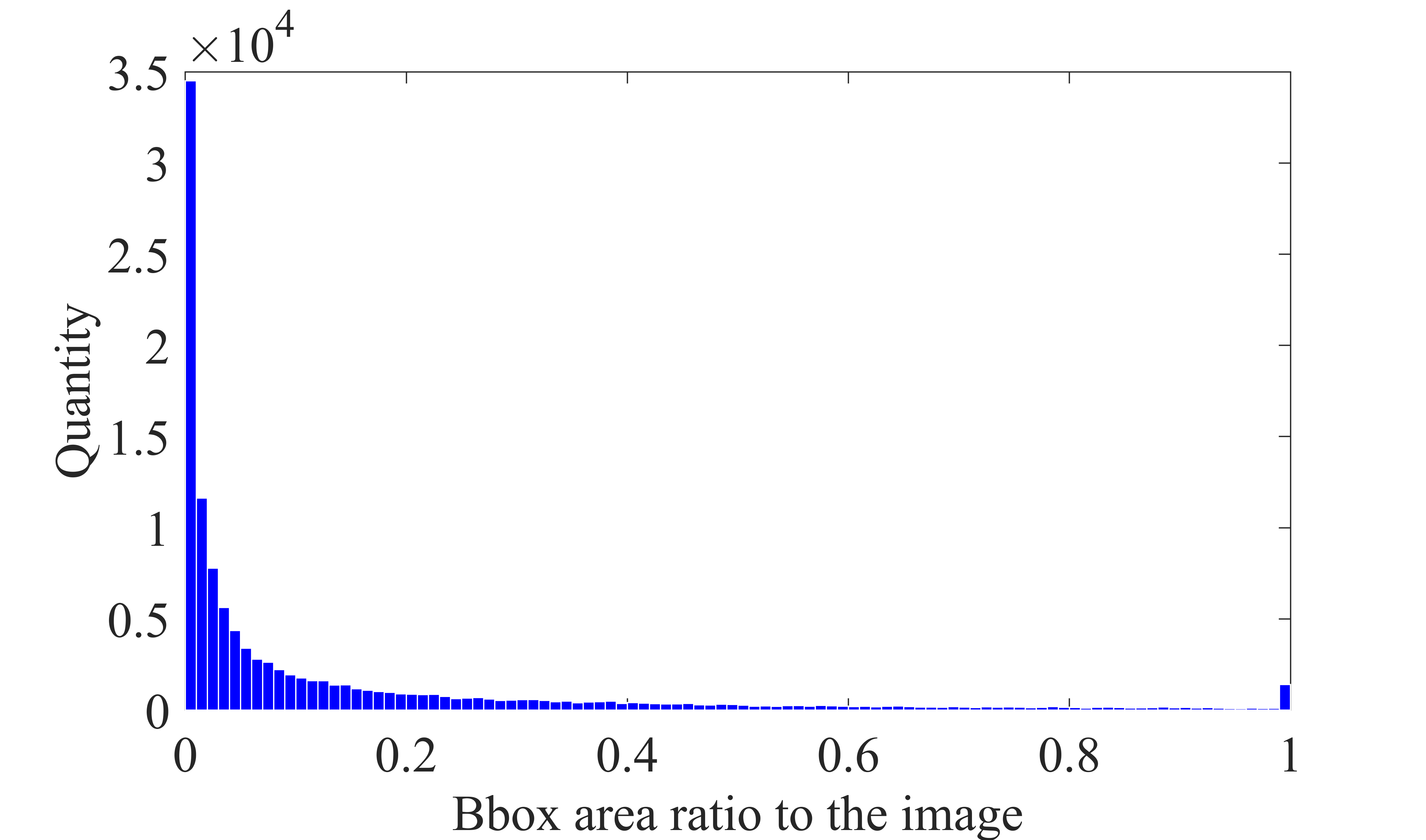}
 \caption{BBox areas and their numbers}\label{fig:area_distribution}
\end{figure}

Besides, the images in OIDV5 have large-scale variation.  Figure \ref{fig:area_distribution} shows histogram of bbox areas and their numbers. Most boxes only account for 10\% of the images, however some boxes hold the whole images. So the scale of annotated boxes is extremely imbalanced.

The most distinct characteristic of OIDV5 is category imbalance. The largest class has hundreds of thousands of bounding boxes, and the smallest class has only a dozen. If using all images and its bounding boxes, it will take dozens of days to converge. 

\begin{figure*}[htbp]
 \centering
 \includegraphics[width=14cm]{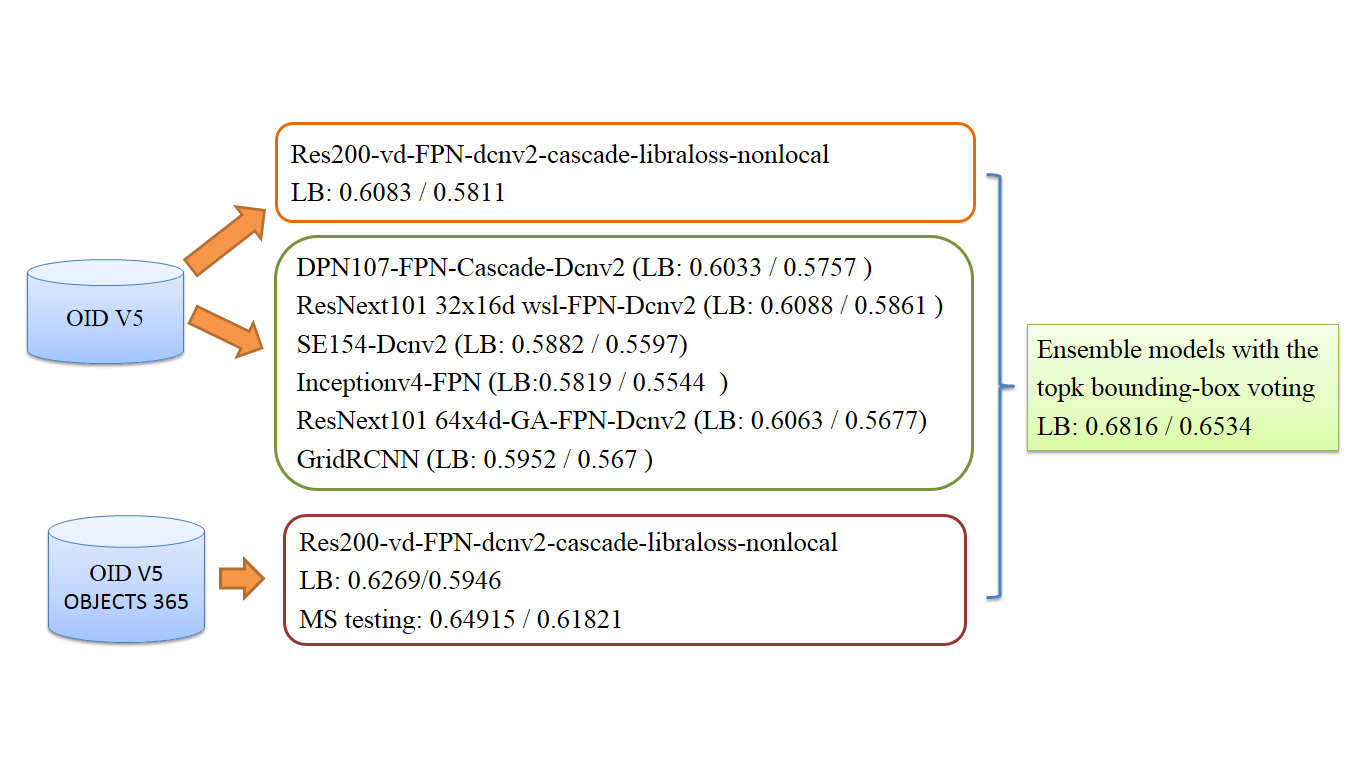}
 \caption{Framework of our solution.}\label{fig:all_liucheng}
\end{figure*}

Figure \ref{fig:all_liucheng} shows the framework of our solution. It contains four key points: (1) we train an extraordinary single model with Res200-vd ~\cite{DBLP:journals/corr/HeZRS15,DBLP:journals/corr/abs-1812-01187}, FPN ~\cite{DBLP:journals/corr/LinDGHHB16}, Dcnv2~\cite{DBLP:journals/corr/abs-1811-11168}, Cascade ~\cite{cai_cascade_2017}, Non-local ~\cite{DBLP:journals/corr/abs-1711-07971} and libra loss ~\cite{DBLP:journals/corr/abs-1904-02701} based on PaddlePaddle. (2) Since the 189 classes are same between OIDV5 and Objects365, we combine these two datasets to enhance the ability of the above model. (3) To further promote the performance, the detectors with the different backbones and architectures are trained. (4) We propose a modified version of NMS to ensemble models. The ensemble box is obtained by the top-k bounding-boxes voting. It will make better use of the above trained models.

%------------------------------------------------------------------------
\section{Baseline}

\begin{figure}[htbp]
 \centering
 \includegraphics[width=7cm]{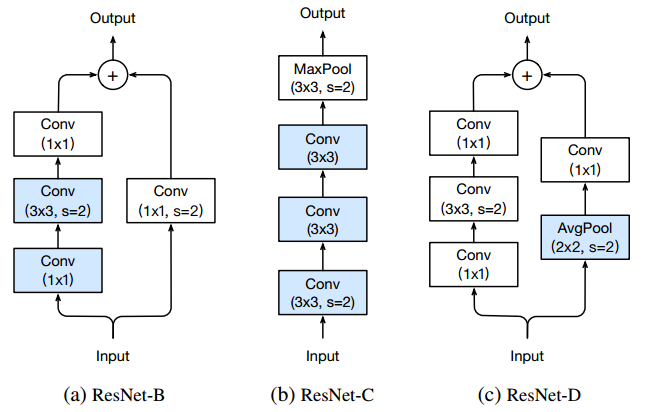}
 \caption{Different ResNet structures.(From left to right, they are ResNet-vb, ResNet-vc and ResNet-vd)}\label{fig:resnet-vbcd}
\end{figure}

We mainly leverage Faster RCNN ~\cite{DBLP:journals/corr/RenHG015} framework for the object detection task. Backbone is vital to the model performance, especially for large datasets. ResNet200-vd is utilized as backbone for our baseline model. ResNet-vb, ResNet-vc and ResNet-vd structures are shown as Figure \ref{fig:resnet-vbcd}.

The pretrained model is based on ImageNet-1k. In the training phase, we used the mixup and label smoothing strategies to avoid overfitting and enhance the robustness of the model. In the end, the top-1 accuracy of ResNet200-vd on ImageNet-1k is 80.93\%. It is worth mentioning that ResNet-vd performs better than ResNet-vb generally in classification and object detection task. The top-1 accuracy of ResNet50-vb and ResNet50-vd are 76.50\% and 79.12\%, respectively.

Feature Pyramid Networks is vital to the model performance, especially for small objects detection. Cascade RCNN can help us to obtain a detector with higher quality. Deformable ConvNets V2 has the ability to focus on pertinent image regions. Non-local Network can aggregate feature map context information. Compared to smooth-l1 loss, libra loss has the advantage of balancing the losses of inliers and outliers. We train a detector using all the above methods based on PaddlePaddle. To validate the model's effectiveness, we trained the same model on COCO dataset. The final mAP on COCO minival 2017 set reached 51.3\%. 

OIDV5 dataset has serious problem of class imbalance. Therefore, we adopt dynamic sampling strategy during training. For better performance on the various sizes of objects, we perform multi-scale training. To improve recall, we use SoftNMS ~\cite{DBLP:journals/corr/BodlaSCD17} rather than NMS for the detector post processing. If not specified, all the models are evaluated using single scale with SoftNMS.

For the baseline model, we utilize 8 NVIDIA Tesla V100 cards to train our models. What’s more, piecewise decay strategy is adopted for the learning rate schedules. Learning rate is set as 0.01 and reduced at 10$^6$th and 1.4x10$^6$th minibatch with the total iteration number set as 1.5x10$^6$. 

The final scores on public and private leaderboard are shown in table \ref{tab:resnet200_vd_models}. To validate the backbone performance, the ResNet200-vd-base and ResNet200-vd-FPN detector based on Faster RCNN model are also trained. ResNet200-vd-FPN also performs better. Here ResNet200-vd-All denotes the detector is trained using FPN, Cascade, Dcnv2, Non-local, libra loss based on ResNet200-vd backbone.

\begin{table}
\begin{center}
\begin{tabular}{|l|c|}
\hline
Model           & Score(public/private) \\
\hline\hline
    ResNet200-vd-base      & 0.5860/0.5563 \\
    ResNet200-vd-FPN       & 0.5925/0.5666 \\
    ResNet200-vd-All       & 0.6083/0.5811 \\
\hline
\end{tabular}
\end{center}
\caption{Trained models based on ResNet200-vd.}\label{tab:resnet200_vd_models}
\end{table}

%-------------------------------------------------------------------------
\section{Application of Objects365 Dataset}

Objects365 is a brand new dataset for object detection research and it has 365 classes of objects, We find that about 189 classes between OIDV5 and Objects365 dataset are same. This helps us expand dataset more easily. We use the Objects365 dataset in two ways.

\subsection{Pretrained model}

We train a ResNet200-vd-FPN model based on Objects365 Dataset, then use the trained model as pretrained model to train a ResNet200-vd-All model for OIDV5. It’s noted that the model is just trained 10$^6$ minibatches because of the time limit. As can be seen in Tabel \ref{tab:obj365_res}, the model score on public and private leaderboard is 0.60, 0.57076, respectively, which are lower than ResNet200-vd-All model.

\begin{table}
\begin{center}
\begin{tabular}{|l|c|}
\hline
Model           & Score \\
\hline\hline
    Pretrained Objects365 Dataset model      & 0.6000/0.5708 \\
    Trained with Joint Data  & 0.6269/0.5946 \\
   	Trained with Joint Data(MS-testing)      	  & 0.6492/0.6182 \\
    All models using ResNet200-vd(voting)  	  & 0.6582/0.6287 \\
\hline
\end{tabular}
\end{center}
\caption{Results based on Objects365 Dataset.}\label{tab:obj365_res}
\end{table}

\subsection{Train Models with Joint Data}

We use category mapping for the Objects365 dataset. In detail, we expand OIDV5 using Objects365 dataset for the duplicate categories. The final model score on public and private leaderboard are 0.6269 and 0.59459. We believe that the huge improvement is attributed to more data and more detailed annotation information.

Score of multi-scale test result is 0.64915 and 0.61821. The significant improvement compared to singe scale result is mainly credited with our merge strategy, which will be introduced later.

We merge all the model results based on ResNet200-vd backbone, and the final scores on public and private leaderboard are 0.65821 and 0.62874.

%-------------------------------------------------------------------------
\section{The Ensemble of Models}

\subsection{Top-k Voting-nms}

\begin{figure}[htbp]
 \centering
 \includegraphics[width=8cm]{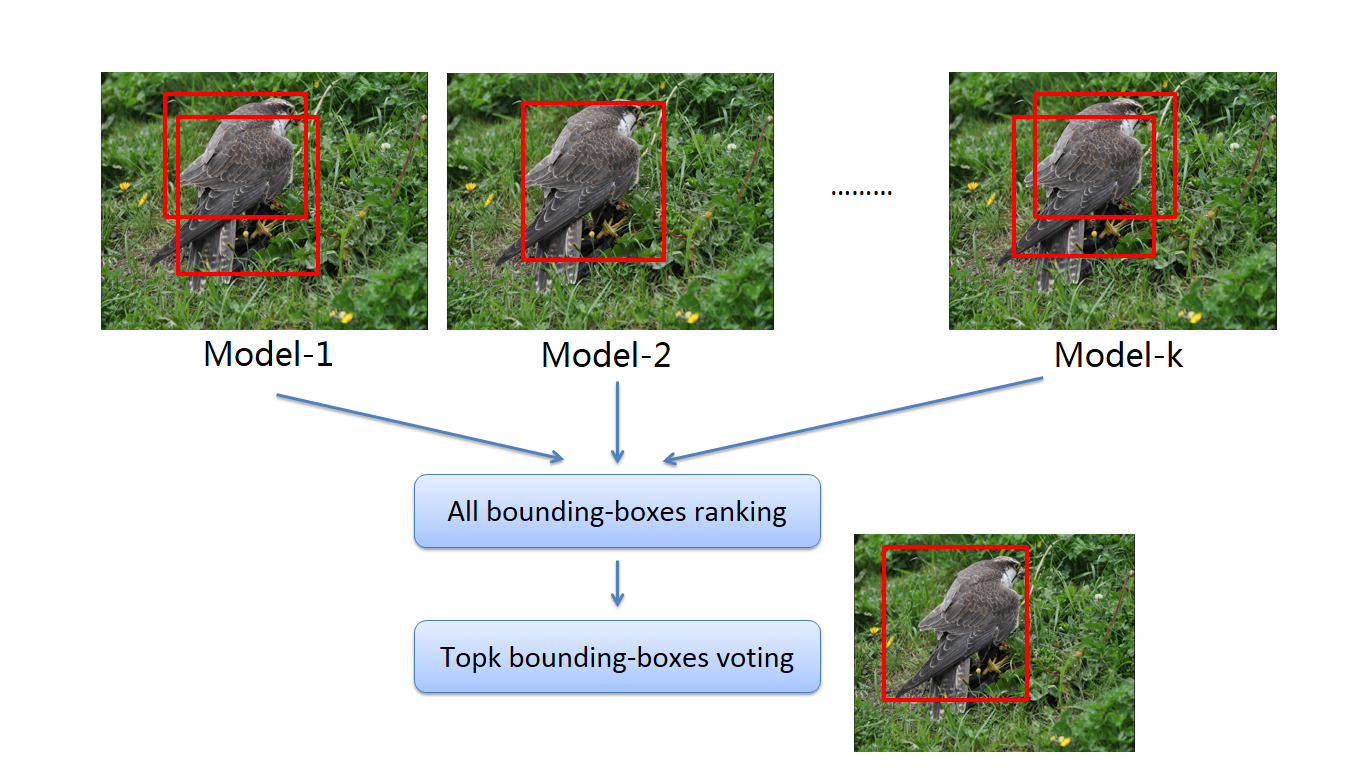}
 \caption{Top-k bounding-box voting strategy}\label{fig:voting-s}
\end{figure}

Intuitively, the results of different models has different character, bounding box ensemble will be beneficial to the detection performance. For each model, since we filtered the bounding boxes based on soft-nms, however, it’s helpless to reduce the duplicated boxes directly with same label in the same location. In this paper, Top-k voting-nms method is introduced based on voting mechanism. Based on the results of different models, we consider the average of the top-k bounding-boxes’ scores and locations as the predicted score and location. Instead of keeping the highest one only in the original NMS, the ensemble box is obtained by the top-k bounding-boxes voting. By this way, the bounding boxes with same label are filtered by voting strategy. The IOU threshold is 0.5 in this step. Figure \ref{fig:voting-s} shows the procedure voting strategy in this paper.

To illustrate the power of the top-k voting-nms, the multi-scale testing (ms testing) of the model trained from the OIDV5 and Objects365 dataset combine dataset (LB 0.6269/0.5946). Using the original nms, the mAP of the ms testing result is 0.5926 in private dataset which is not promoted. However, the top-k voting-nms with only refining the score has a better result, the mAP is achieved to 0.6131. Further, when both refine score and location is adopted, the mAP is improved to 0.6492/0.6182.

\begin{table}
\begin{center}
\begin{tabular}{|l|c|c|}
\hline
Method & Score \\
\hline\hline
    Soft-nms    & 0.6269/0.5946 \\
    Voting soft-nms    & 0.6271/0.5926 \\
    Top-k voting-nms(score)    & 0.6430/0.6131 \\
    Top-k voting-nms(score and location)	    & 0.6492/0.6182 \\
    
\hline
\end{tabular}
\end{center}
\caption{Trained models for ensemble.}\label{tab:other_models}
\end{table}

%-------------------------------------------------------------------------
\subsection{Model Ensemble Result}

Infactly, different model architectures have different perception on image. The models trained on OIDV5 dataset are ensembled to improve the performance. Specifically, the other models and scores are listed as Table \ref{tab:other_models}. Here, the models are just trained sixty thousand minibatches. All the models are trained with corresponding FPN pretrained models.

\begin{table}
\begin{center}
\begin{tabular}{|l|c|c|}
\hline
Model           & Score \\
\hline\hline
	All models using ResNet200-vd(voting) & 0.6582/0.6287 \\
    DPN107-FPN-Cascade-Dcnv2      & 0.6033/0.5757 \\
    ResNext101\_32x16d\_wsl-FPN-Dcnv2 & 0.6088/0.5861 \\
    SE154-Dcnv2 ~\cite{hu_squeeze-and-excitation_2017}     			  & 0.5882/0.5597 \\
    Inceptionv4-FPN  			  & 0.5819/0.5544 \\
    ResNext101\_64x4d-GA-FPN-Dcnv2 & 0.6063/0.5677 \\
    GridRCNN-GN    & 0.5952/0.5670 \\
    Voting results & 0.6816/0.6534 \\
 
\hline
\end{tabular}
\end{center}
\caption{Trained models for ensemble.}\label{tab:other_models}
\end{table}

Besides classic methods, some novel object detection frameworks are adopted. In ResNext101\_64x4d-GA-FPN-Dcnv2 ~\cite{DBLP:journals/corr/abs-1901-03278}, ResNext101\_64x4d is used as backbone, and guided anchoring ~\cite{DBLP:journals/corr/abs-1901-03278} is adopted to produce region proposal. In GridRCNN ~\cite{DBLP:journals/corr/abs-1811-12030}, which is a novel object detection framework, to take the full advantage of the correlation of points in a grid, it adopts a grid guided localization mechanism for accurate object detection. We present detailed benchmarking results for some methods in Table \ref{tab:other_models}. The scores after merging all the above models are 0.6816 and 0.6534 on public and private leaderboard.

\section{Conclusion}

In this paper, we present an object detection framework based on PaddlePaddle, which performs well not only on COCO dataset but also on OIDV5. A new top-k voting method is proposed for model ensemble, which helps us win 2nd place in Google AI Open Images Object Detection Track 2019.

{\small
\bibliographystyle{ieee_fullname}
\bibliography{egbib}
}

\end{document}